\documentclass[10pt,twocolumn,letterpaper]{article}

\usepackage[pagenumbers]{cvpr}             

\usepackage{graphicx}
\usepackage{amsmath}
\usepackage{amssymb}
\usepackage{booktabs}
\usepackage[normalem]{ulem}
\usepackage{stfloats}
\usepackage{balance}

\usepackage[usenames,dvipsnames]{xcolor}
\graphicspath{ {./figures/} }

\usepackage[pagebackref,breaklinks,colorlinks]{hyperref}
\usepackage{verbatim}

\usepackage[capitalize]{cleveref}
\crefname{section}{Sec.}{Secs.}
\Crefname{section}{Section}{Sections}
\Crefname{table}{Table}{Tables}
\crefname{table}{Tab.}{Tabs.}

\def\cvprPaperID{1657} 
\def\confName{CVPR}
\def\confYear{2022}
\providetoggle{showcomments}
\settoggle{showcomments}{true} 								%

\iftoggle{showcomments}{
    
    \newcommand{\todo}[1]{\textcolor{blue}{\textbf{TODO:} #1}}
    \newcommand{\resolved}[3][]{\ifstrequal{#1}{resolved}{\textcolor{blue}{RESOLVED:}~\textbf{{\MakeUppercase #2:}}~{#3}}{\textbf{\MakeUppercase #2:}~#3}}
    \newcommand{\andrea}[2][]{\textcolor{ForestGreen}{\resolved[#1]{andrea}{#2}}}
    \newcommand{\vitto}[2][]{\textcolor{red}{\resolved[#1]{vitto}{#2}}}
    \newcommand{\jasper}[2][]{\textcolor{violet}{\resolved[#1]{jasper}{#2}}}
    \newcommand{\tm}[2][]{\textcolor{magenta}{\resolved[#1]{TM}{#2}}}
    
}{
    
    \newcommand{\todo}[1]{}
    \newcommand{\andrea}[2][]{}
    \newcommand{\vitto}[2][]{}
    \newcommand{\jasper}[2][]{}
    \newcommand{\tm}[2][]{}
    
}

\newcommand{\softiou}{SoftIoU-EEP}
\newcommand{\msleep}{MS-LEEP}
\newcommand{\ioueep}{IoU-EEP}
\newcommand{\eleep}{E-LEEP}
\newcommand{\para}[1]{\par\noindent\textbf{#1}}
\begin{document}
\title{Transferability Metrics for Selecting Source Model Ensembles}

%
%


\author{
Andrea Agostinelli \quad 
Jasper Uijlings \quad
Thomas Mensink \quad
Vittorio Ferrari \\[1mm]
Google Research\\
{\tt\small \{agostinelli, jrru, mensink, vittoferrari\}@google.com}

\vspace{1.1cm}
}

\maketitle

\begin{abstract}
   We address the problem of ensemble selection in transfer learning: Given a large pool of source models we want to select an ensemble of models which, after fine-tuning on the target training set, yields the best performance on the target test set. Since fine-tuning all possible ensembles is computationally prohibitive, we aim at predicting performance on the target dataset using a computationally efficient transferability metric.
   We propose several new transferability metrics designed for this task and evaluate them in a challenging and realistic transfer learning setup for semantic segmentation:
   we create a large and diverse pool of source models by considering 17 source datasets covering a wide variety of image domain, two different architectures, and two pre-training schemes.
   Given this pool, we then automatically select a subset to form an ensemble performing well on a given target dataset.
   We compare the ensemble selected by our method to two baselines which select a single source model, either (1) from the same pool as our method; or (2) from a pool containing large source models, each with similar capacity as an ensemble.
   Averaged over 17 target datasets, we outperform these baselines by 6.0\% and 2.5\% relative mean IoU, respectively.
   
\end{abstract}

\section{Introduction}
\label{sec:intro}

In transfer learning we want to re-use knowledge previously learned on a source task to help learning a target task. The most common form of transfer learning in computer vision is to pre-train a single source model on the generic ILSVRC dataset~\cite{azizpour15pami, chu16eccv, girshick15iccv, he17iccv, huh16neurips, kornblith19cvpr, shelhamer16pami, zhou19arxiv} and then fine-tune it on the target dataset. However, often a more domain-specific approach can lead to better results \cite{mensink21arxiv, ngiam18arxiv, yan20cvpr}.
Hence, it is beneficial to have a large pool of diverse source models such that it contains models suited for many different target tasks. The problem then becomes: how can we automatically and efficiently select good source models for a given target task?

Recently, transferability metrics were introduced to address this problem~\cite{bao19icip, nguyen20icml, li21cvpr, you21icml, tan21cvpr, tran19iccv}. The general goal is to select a single source model which, after fine-tuning on the target training set, yields the best performance on the target test set. Transferability metrics enable to select this model efficiently without carrying out expensive fine-tuning on the target training set.

While previous transferability metrics consider selecting a single source model, in this paper we aim at selecting a subset containing multiple source models to form an {\em ensemble}.
Ensembles are a general technique used to improve model accuracy, out-of-distribution robustness, and to estimate uncertainty \cite{breiman96ml, dietterich00springer, freund96icml, fort19arxiv, krogh95nips, lakshminarayanan17neurips, ovadia19nips, wang20arxiv}.
Furthermore, by aggregating multiple source models, we can combine knowledge coming from multiple source datasets and image domains, which may be beneficial for a particular target task.
Hence, in this paper we extend previous work on transferability by proposing several transferability metrics designed for {\em ensemble selection}.

To evaluate ensemble selection we introduce a challenging experimental setup.
We consider semantic segmentation as a task, with a truly diverse pool of source models, as we train them on 17 complete datasets spanning a wide variety of images domains, while also varying their model architectures and pre-training schemes (Fig.~\ref{fig:pipeline}).
In contrast, previous works typically
focus on image classification~\cite{bao19icip, nguyen20icml, li21cvpr, you21icml, tan21cvpr, tran19iccv},
consider a narrower range of at most 4 source datasets \cite{tan21cvpr,nguyen20icml,you21icml,li21cvpr},
and often generate multiple datasets artificially by sampling different subsets of classes out of a single actual dataset~\cite{bao19icip, nguyen20icml, tan21cvpr, tran19iccv}.

To summarize, we make the following contributions:
(1) We design transferability metrics for \emph{ensemble} selection.
(2) We consider a challenging application scenario on semantic segmentation featuring a large and truly diverse pool of source models.
(3) We compare the ensemble selected by our method to two baselines which select a single source model, either from the same pool as our method; or from a pool containing large source models, each with similar capacity as an ensemble.
Averaged over 17 target datasets, we outperform these baselines by 6.0\% and 2.5\% relative mean IoU, respectively (Sec. \ref{sec:ensemble_selection_results}).

\section{Related Work} \label{related_work}
\label{sec:related-works}

\para{Transfer Learning.}
The most common form of transfer learning in computer vision is to pre-train a model on ILSVRC'12~\cite{dosovitskiy20arxiv, kolesnikov20eccv}, and fine-tune it on the target dataset.
Several works extend this to using a larger source dataset such as ImageNet21k (9M images), JFT-300M (300M images) \cite{dosovitskiy20arxiv, kolesnikov20eccv, mustafa20arxiv}, or Open Images (1.7M images) \cite{yan20cvpr}.
Other works consider self-supervised pre-training, enabling the use of unlabeled source datasets (\eg~\cite{he20cvpr, chen21cvpr, chen20icml, islam21iccv, li21iccv}).

Several studies explore in-depth under which circumstances transfer learning works.
Mensink et al.~\cite{mensink21arxiv} study transfer learning across datasets with vastly different image domains and multiple visual tasks.
Mustafa et al.~\cite{mustafa21arxiv} studies transfer learning in medical imaging.
Finally, Taskonomy~\cite{zamir18cvpr} establishes relationships between visual tasks (\eg semantic segmentation, depth prediction, etc.).
Following the success of Taskonomy, several works investigate whether visual task relatedness can be predicted~\cite{dwivedi19cvpr, song19neurips, song20cvpr, dwivedi20eccv, bolya21neurips} rather than calculated by brute force~\cite{zamir18cvpr}.



Only a few works conduct transfer learning from multiple source datasets at the same time.
Liu et al.~\cite{liu19iclr} train a student model using knowledge from
multiple teachers (\ie source models), which is expensive in both memory and computation.
Zoo-Tuning \cite{shu21icml} learns to aggregate the parameters of multiple source models into a target model. This requires storing all source models in memory during training, limiting scalability.
In contrast, we select an ensemble from a large pool of source models in a computationally and memory efficient manner.

\para{Transferability Metrics.}
Recently, several papers introduced transferability metrics.
H-score \cite{bao19icip} measures the discriminativeness of source model features on the target task in terms of inter-class and intra-class variance. 
LEEP~\cite{nguyen20icml} measures how well a classifier built on top of source model predictions performs on the target task.
$\mathcal{N}$LEEP~\cite{li21cvpr} trains a Gaussian Mixture Model (GMM) on top of source model features. Then it measures how well a classifier built on top of these GMM predictions performs on the target task. 
LogME~\cite{you21icml} estimates accuracy on the target task based on a formulation which integrates over all possible linear classifiers built on top of the source model features.
OTCE~\cite{tan21cvpr} applies a source model to extract image features from both the source and target dataset. Then it uses optimal transport between these features to calculate domain difference and task difference. 
Finally, NCE~\cite{tran19iccv} considers a more restrictive setting where the source and target datasets consist of identical images. Their method uses conditional entropy between ground truth source and target labels, which avoids training models and is thus computationally efficient.

To put our work in context:
(1) Instead of selecting a single source model~\cite{bao19icip, nguyen20icml, li21cvpr, you21icml, tan21cvpr}, we do \emph{ensemble} selection.
(2) Instead of image classification~\cite{bao19icip, nguyen20icml, li21cvpr, you21icml, tan21cvpr, tran19iccv}, we address semantic segmentation.
(3) We consider a larger variety of source datasets than previous works (17 vs at most  4~\cite{tan21cvpr}).
(4) We consider complete datasets, whereas previous works often sample different subsets of classes out of a single actual dataset~\cite{bao19icip, nguyen20icml, tan21cvpr, tran19iccv}.


\para{Ensemble of Models.} \label{para:ensemble}
Ensembling machine learning models is a classical method for increasing accuracy~\cite{breiman96ml,dietterich00springer,freund96icml,hansen90pami,krogh95nips,lee15arxiv}, where having diverse models is typically important. More recently, ensembles of deep neural network have been studied in the context of uncertainty estimation and out-of-distribution robustness~\cite{allingham21arxiv, fort19arxiv, lakshminarayanan17neurips, ovadia19nips}.
\section{Methods} \label{sec:methods}

We consider the problem of source model ensemble selection for semantic segmentation.
Given $N$ source models and a target dataset, the goal is to select an \emph{ensemble} of source models which, after fine-tuning on the target training set, yields the best performance on the target test set. Since fine-tuning all possible ensembles is too computationally expensive, we predict performance on the target dataset using a computationally efficient transferability metric.

\cref{preparing_source_models} discusses what makes for a good pool of source models and describes how we construct this pool. 
\cref{metric-semantic-segmentation} describes our setup to work with transferability metrics in semantic segmentation tasks.
\cref{leep-single} describes LEEP~\cite{nguyen20icml}, a transferability metric for single-source selection. 
We use this as a starting point in~\cref{metric-multi-source} to define our four transferability metrics for \emph{ensemble} selection.


\subsection{Preparing source models} \label{preparing_source_models}

We want to create a pool with a large variety of source models for three reasons:
(1) this increases the chance that for any given target dataset there exists at least one good source model.
(2) we need bad source models to verify that our transferability metrics correctly select good source models while discarding bad ones.
(3) an ensemble can only outperform its individual members if they are diverse (and therefore complementary)~\cite{bian21ieee, dietterich00springer, freund96icml, hansen90pami, krogh95nips}.
Hence, we construct our source model pool by incorporating diversity in three ways: we use 17 different source datasets, two model architectures and two pre-training strategies.
Tab.~\ref{tab:setup_comparison} summarizes how this setup compares to related work.

\begin{table}[t!]
\footnotesize	
\vspace{-0.6cm}
  \centering
  \setlength\tabcolsep{2.8pt}
  \begin{tabular}{l | c c c c c c}
    \toprule
     & Ours & Ours & LEEP & LogME & OTCE & $\mathcal{N}$LEEP \\
     & \cref{sec:ensemble_selection} & \cref{sec:evaluate-metrics} & \cite{nguyen20icml}
     & \cite{you21icml} & \cite{tan21cvpr} & \cite{li21cvpr} \\
    \midrule
     \# source datasets & 17 & 10-15 & 1 & 1 & 4 & 3\\
     \# pre-training schemes & 2 & 1-2 & 1 & 1 & 1 & 4\\
     \# model architectures & 2 & 1-2 & 9 & 10 & 1 & 13\\
     \# source models & 68 & 15 & 9 & 10 & 4 & 41\\
    \midrule
    \# candidates & 41K & 455 & 9 & 10 & 4 & 41 \\
  \end{tabular}
  \caption{Comparing our experimental setup (see Sec. \ref{sec:evaluate-metrics} and \ref{sec:ensemble_selection}) to previous works on cross-dataset source selection. We compare the diversity of source models in terms of the number of source datasets, pre-training schemes, and model architectures.
  The last row denotes the number of candidate source models (or ensembles in our case) that are in the pool for a given target dataset. For~\cite{li21cvpr, nguyen20icml, tan21cvpr, you21icml} we consider their largest source selection experiment.}
  \label{tab:setup_comparison}
  \vspace{-0.4cm}
\end{table}

\para{Source datasets.}
The image domain is one of the most important factors to influence whether transfer learning will succeed~\cite{ngiam18arxiv, mensink21arxiv, pan09ieee}, and therefore we want to cover a wide array of image domains. Furthermore, the most natural way to perform transfer learning is to consider each dataset as a whole (rather than subsampling a dataset to simulate dataset variations~\cite{bao19icip, nguyen20icml, tan21cvpr, tran19iccv}). 
Therefore, we adopt the realistic cross-dataset transfer learning setup for semantic segmentation by~\cite{mensink21arxiv}: 17 source datasets from 6 image domains (consumer photos, driving, aerial, indoor, underwater, synthetic; Tab.~\ref{tab:datasets}).
While this setup was defined in~\cite{mensink21arxiv}, that work did not explore any transferability metric.

\para{Model architectures.}
We consider two semantic segmentation architectures, each with a backbone and a linear classification layer.
As the first backbone we choose HRNetV2~\cite{wang20pami}, a high-resolution alternative to ResNet. It maintains parallel feature representations at different resolutions, which helps dense prediction tasks~\cite{lambert20cvpr,mensink21arxiv,wang20pami}. As ensembles contain multiple models, we choose a light-weight version: HRNetV2-W28 (23M parameters).

\begin{table}[pt]
\vspace{-0.6cm}
    \centering
    \resizebox{0.5\textwidth}{!}{
        \begin{tabular}{llcc}\toprule
        Dataset                      & Domain  & \# classes  & \# train images  \\
        \midrule
        Pascal Context~\cite{mottaghi14cvpr}  & Consumer              & 60        & 5K \\
        Pascal VOC~\cite{pascal-voc-2012} & Consumer              & 22   &  10K \\
        ADE20K~\cite{zhou17cvpr}      & Consumer              & 150   &   20K \\
        COCO Panoptic~\cite{caesar18cvpr, lin14eccv, kirillov19cvpr}       & Consumer              & 134 & 118K \\
        KITTI~\cite{alhaija18ijcv}   & Driving & 30 & 150 \\
        CamVid~\cite{brostow09prl}    & Driving  & 23 & 367 \\
        CityScapes~\cite{cordts16cvpr}    & Driving  & 33   & 3K \\
        India Driving Dataset (IDD)~\cite{varma19wacv} & Driving & 35 & 7K \\
        Berkeley Deep Drive (BDD)~\cite{yu20cvpr} & Driving & 20 & 7K \\
        Mapillary Vista Dataset~\cite{neuhold17cvpr} & Driving & 66 & 18K \\
        ISPRS~\cite{rottensteiner14isprs}  & Aerial & 6 & 4K \\
        iSAID~\cite{waqas2019isaid, xia18cvpr} & Aerial     & 16 & 27K \\
        SUN RGB-D~\cite{song15cvpr} & Indoor & 37 & 5K \\
        ScanNet~\cite{dai17cvpr} & Indoor & 41 & 19K \\
        SUIM~\cite{islam2020suim} & Underwater & 8 & 1525 \\
        vKITTI2~\cite{cabon20arxiv, gaidon16cvpr} & Synthetic driving & 9 & 43K \\
        vGallery~\cite{weinzaepfel19cvpr} & Synthetic indoor & 8 & 44K \\
        \bottomrule
        \end{tabular}
    }
    \caption{Semantic segmentation datasets used in our paper.}
    \label{tab:datasets}
    \vspace{-.4cm}
\end{table}

As the second backbone, we adopt a high-resolution variant of ResNet50~\cite{he16cvpr}.
First, we remove the downsampling operations in the last two ResNet blocks while increasing the dilation rate~\cite{wu19pr}.
Second, we add an upsampling layer using 5 parallel atrous convolutions with different dilation rates, which enlarges the field of view of the filters without compromising on spatial resolution \cite{chen17pami}.
Finally, we remove the last four layers of the last ResNet block to make this backbone have the same number of parameters as HRNetV2-W28 (we call this model ResNet23M).
As all ensembles we build contain the same number of source models, this ensures that they also have the same number of parameters, enabling a fair comparison between them.


\para{Pre-training schemes.}
Fully supervised pre-training on ILSVRC'12 generally benefits semantic segmentation \cite{mensink21arxiv,shelhamer16pami}.
Furthermore, self-supervised pre-training is making rapid progress and can even outperform fully supervised pre-training~\cite{li21iccv,islam21iccv}.
To maximize model diversity, we create two variants of each source model by using two types of ILSVRC'12 pre-trained weights: fully supervised and using the self-supervised SimCLR method~\cite{chen20icml}.


\para{Training source models.} \label{para:training-setup}
We have 17 source datasets, two architectures, and two pre-training schemes.
We train a source model for each combination, i.e. 68 in total (details in the Appendix B). These models are trained only once and reused in all experiments.


\subsection{Setting up for semantic segmentation} \label{metric-semantic-segmentation}

In most previous works \cite{you21icml, nguyen20icml, tan21cvpr, bao19icip, tran19iccv, li21cvpr}, a transferability metric is primarily applied to image classification, where an image is associated to one label.
In semantic segmentation instead we have predictions at the pixel level,
and therefore we consider for each pixel $x_i$ and its ground-truth label $y_i$ as an individual example $(x_i, y_i)$.

The number of datapoints in semantic segmentation is approximately 6 orders of magnitude higher than in image classification. To reduce the computational cost, we sample 1000 pixels per image to calculate each transferability metric. 
%
Furthermore, semantic segmentation datasets often have large class imbalance, which can negatively affect results.
Therefore, we sample pixels inversely proportionally to the frequency of their class labels in the target dataset.

\subsection{LEEP as single-source transferability metric}\label{leep-single}

We want a transferability metric suitable for selecting an ensemble of models for semantic segmentation.
We start from LEEP~\cite{nguyen20icml}, which is based on probability distributions, for several reasons.
First, both per-pixel predictions and ensemble selection increase computational and memory complexity compared to the usual image classification and single-source selection.
LEEP is computationally cheap, requiring just a single forward pass of the target training set through the source model, without additional training. Second, LEEP is memory-efficient as it stores predictions instead of features as opposed to alternative metrics~\cite{you21icml, tan21cvpr, bao19icip}.
Finally, starting from LEEP we can derive clear mathematical formulations for the multi-source setting.

LEEP calculates a transferability score between a single source model $s$ and a target training set $\mathcal{D}_t$, containing a set of training samples $(x_i, y_i)\in\mathcal{D}_t$.
Applying $s$ to a target sample produces the probability $p_{\textrm{s}}(z|x_i)$, for each source class $z$ in the source label space $\mathcal{Z}$.
The core idea of LEEP is to associate predictions in the source label space $\mathcal{Z}$ to predictions in the target label space $\mathcal{Y}$.
To do so, we apply $s$ to all samples in $\mathcal{D}_t$, and then compute the empirical joint distribution $\hat{P}(y,z)$ measuring co-occurrences between all pairs of labels $(y,z) \in \mathcal{Y} \times \mathcal{Z}$. 

Next, we can calculate the empirical conditional distribution $\hat{P}(y|z)$ as $\frac{\hat{P}(y,z)}{\hat{P}(z)}$.
Given the source model $s$, we can now construct a classifier, called  \textbf{E}xpected \textbf{E}mpirical \textbf{P}redictor (EEP): 
\begin{equation}
  p_s(y_i|x_i) = \sum_{z\in \mathcal{Z}}{\hat{P}(y_i|z) p_{\textrm{s}}(z|x_i)}
  \label{eq:eep}
\end{equation}
Here $p_s(y_i|x_i)$ is the probability that model $s$ assigns to the ground-truth label $y_i$ at pixel $x_i$.
LEEP is defined as the log-average of the predictor over $\mathcal{D}_t$:
\begin{equation}
  \mathrm{LEEP} = \frac{1}{n}\sum_{i=1}^{n}\log p_s(y_i|x_i)
  \label{eq:leep}
\end{equation}
We can see that LEEP measures how well the constructed classifier EEP performs on $\mathcal{D}_t$, where better transferability is associated with higher LEEP scores.


\subsection{Multi-source selection transferability metrics} \label{metric-multi-source}

We design four transferability metrics suitable for multi-source selection.
We base all our approaches on the EEP predictor \eqref{eq:eep}, which provides a mathematical foundation to establish relationships between source models.
In all cases our ensembles contain a fixed number $S$ of source models.

\para{Multi-Source LEEP (MS-LEEP).}
A natural way of extending LEEP to the multi-source setting is to compute the joint probability distribution over the $S$ source model predictions, where each model makes predictions in its own label space.
%
%
%
But this requires calculating the joint probability distribution for every possible subset of $S$ models, which is infeasible for a large pool of $N$ models.
Instead, we assume the source models to be independent, yielding a simplified joint conditional distribution:
\begin{equation}
    \hat{P}(y|z_1,z_2,...,z_S) \approx  \prod_{s=1}^{S}{\hat{P}(y|z_s)}
    \label{eq:proxy_multi_source_cond_distribution}
\end{equation}
We extend \eqref{eq:leep} by applying \eqref{eq:proxy_multi_source_cond_distribution} to define a new metric:
\begin{equation}
    \begin{split}
    \mathrm{MS{-}LEEP} & = \frac{1}{n} \sum_{i=1}^{n} \log \left( \prod_{s=1}^{S}\left({\sum_{z_s\in \mathcal{Z}_s}{\hat{P}(y_i|z_s) p_{\textrm{s}}(z_s|x_i)}}\right)\right) \\ & = \frac{1}{n} \sum_{i=1}^{n} \left(\sum_{s=1}^{S}{\log p_s(y_i|x_i)}\right) = \sum_{s=1}^{S}{\mathrm{LEEP}_s}    
        \label{eq:p-leep}
    \end{split}
\end{equation}
%
Hence MS-LEEP can be seen as taking the best source models according to the single-model metric LEEP. This suggests that other existing transferability metrics~\cite{bao19icip, li21cvpr, you21icml, tran19iccv} can be similarly adapted to ensemble selection.

\para{Ensemble LEEP (E-LEEP).}
We now approach the problem from a different prospective, stressing that we want to predict the transfer performance of an {\em ensemble} of models.
For this we consider the ensemble prediction as the average of the $S$ individual models predictions.
Considering \eqref{eq:eep} as a single-source predictor $p_s(y_i|x_i)$, we can construct the prediction of the ensemble as
\begin{equation}
p_{\textrm{ens}}(y_i|x_i) = \frac{1}{S}\sum_{s=1}^{S} p_s(y_i|x_i)
\label{eq:ensemble-predictor}
\end{equation}
By reformulating \eqref{eq:leep} accordingly, we get a new metric:
\begin{equation}
  \mathrm{E{-}LEEP} = \frac{1}{n}\sum_{i=1}^{n}\log p_{\textrm{ens}}(y_i|x_i)
  \label{eq:e-leep}
\end{equation}
The difference with \msleep{} \eqref{eq:p-leep} is the order of the log and the sum: \eleep{} uses the log of the mean predictions, while \msleep{} uses the mean of the log predictions.

\para{IoU-EEP.} \label{iou-eep}
While \msleep{} and \eleep{} estimate pixel classification accuracy, semantic segmentation performance is usually measured as Intersection-over-Union (IoU).
In practice, for IoU we count True Positives (TP), False Positives (FP), False Negatives (FN), and calculate $\frac{TP}{TP + FP + FN}$.
IoU is calculated separately per class, and then averaged into a single metric (mean IoU).
In order to design a transferability metric that more directly approximates mean IoU, we first use the ensemble predictor \eqref{eq:ensemble-predictor} to compute the predicted semantic segmentation over the target training set
\begin{equation}
  y_i^* = \arg\max_{y \in \mathcal{Y}} p_{\textrm{ens}}(y|x_i)
  \label{eq:e_eep_argmax}
\end{equation}
where $y$ now iterates over the entire target label space (as opposed to being the one ground-truth label $y_i$). 
Finally, we use these predictions $y_i^*$ to calculate the mean IoU, arriving at the \ioueep{} transferability metric.
This metric is also less dependent on the probabilistic output of the source classifiers, which are often poorly calibrated \cite{guo17icml} and therefore could negatively affect the metric.

\para{SoftIoU-EEP.} \label{soft-iou-eep}
Applying the $\arg\max$ in Eq.~\eqref{eq:e_eep_argmax} alleviates calibration errors. But it also loses the fine-grained information in the probability distribution $p_{\textrm{ens}}(\cdot|x_i)$, which could be helpful in ranking similar models.
Hence, we propose to introduce a "Soft-IoU" relaxation. For every pixel $x_i$, instead of counting True Positives, we aggregate their confidences $p_{\textrm{ens}}(y_i|x_i)$ (where $y_i$ is the ground-truth label of that pixel). Therefore the higher the confidences of correct predictions, the higher the predicted transferability.
We do the analogue for errors (FP and FN) by aggregating $1 - p_{\textrm{ens}}(y_i|x_i)$. Since errors are in the denominator, the higher their confidences the lower the predicted transferability.
\section{Evaluating transferability metrics} \label{sec:evaluate-metrics}

\subsection{Experimental setup} \label{multi-source-exp-setup}

The standard procedure to evaluate transferability metrics is to consider several candidate source models for a given target dataset, and measure the correlation between
(A) the transferability score and
(B) the actual performance on the target test set after the source model has been fine-tuned on the target training set~\cite{nguyen20icml, li21cvpr, tan21cvpr, you21icml}. 

In this paper we consider multiple candidate source model \emph{ensembles} for a given target dataset. To measure the desired correlation, we calculate transferability as in Sec.~\ref{metric-multi-source} and we also need to calculate the actual performance of each ensemble on the target dataset.

Considering ensembles leads to additional computational challenges. The number of possible ensembles of $S$ source models out of a pool of $N$ is very large
(the binomial $\binom{N}{S}$). Hence computational efficiency is an important requirement in our experimental design.

\para{Measuring actual performance of ensembles efficiently.}
Given a candidate ensemble, we fine-tune each member model individually and apply it individually on the target test set. Then we take the average of their predictions as the ensemble prediction.
Finally, we measure actual ensemble performance as mean IoU.
By considering the predictions of each ensemble member individually, we can reuse them across ensembles. Hence, we only need to run inference for each model on a target test sample {\em once}, regardless of the number of ensembles the model participates in.

While it would also be possible to fine-tune each ensemble as a whole, this is too computationally expensive for this experiment. We do it in a different setting in Sec.~\ref{sec:ensemble_selection}.

\para{Ensemble size.}
We fix the number of models in an ensemble to $S=3$. We found $S=3$ to be a good compromise to benefit from a diverse ensemble, while limiting overall computation (more details in the Appendix C). 

\para{Subsampling candidate source models.}
For a given target dataset, the total number of candidate source models is 64 (68 minus the 4 trained on that target).
We reduce the number of candidates to gain two types of speed-ups.
Firstly, this requires fine-tuning fewer source models on the target training set.
But more importantly, the number of candidate ensembles grows factorially with $N$: for $N=64$ and $S=3$ there are 41k possible ensembles. While this is not a problem for calculating our transferability metrics, evaluating the performance of 41k ensembles on the target test set is computationally prohibitive.
Hence we limit $N=15$ as described below, yielding $455$ candidate ensembles.

For each target dataset we sample 15 source models as follows. As the final goal is to select a high-performing ensemble, we pick 5 good source models for that target. We first compute all our transferability metrics for all ensembles of $3$ source models from the complete pool ($N=64$). We then select the 5 most frequent models in the top-ranked ensembles across all metrics. 
Since we want to evaluate the ability to distinguish between good and bad sources, we include an additional 10 source models at random.

\para{Target datasets.} We consider five target datasets: Cam\-vid~\cite{brostow09prl}, ISPRS~\cite{rottensteiner14isprs}, vKITTI2 \cite{cabon20arxiv}, KITTI \cite{alhaija18ijcv} and Pascal VOC \cite{everingham12pvoc}. 
Since transfer learning is particularly interesting in scenarios with limited training images~\cite{kolesnikov20eccv,mensink21arxiv,mustafa20arxiv}, we follow \cite{mensink21arxiv} and limit each target training set to 150 images.

\para{Correlation measure: weighted Kendall Tau.}
A transferability metric is useful when it can order the candidate source models (or ensembles) according to their actual performance. The exact predicted performance value is less important.
Therefore we use a rank correlation measure as in~\cite{li21cvpr, mensink21arxiv}. Moreover, as in practice we mainly care about selecting a high performing ensemble, we follow~\cite{you21icml} and use a weighted version of Kendall $\tau$ considering top-ranked items more important than low-ranked ones~\cite{vigna15www}.

\para{Baseline transferability metric.}
As there is no previously proposed transferability metric for source model ensembles, we introduce a simple baseline. It is  based on three factors:
(1) source model performance $P_s$ evaluated on a source test set,
(2) source dataset size $N_s$ in terms of number of images, and
(3) source dataset richness measured by the number of source classes $C_c$.
For a candidate ensemble containing $S$ source models, we calculate this baseline as:
$\mathrm{BASE} = \sum_{s=1}^{S}(P_s \times N_s \times C_s)$.
%
%
This baseline is target-agnostic, favours broad source datasets (COCO, ADE20k, Mapillary), and the best model architecture (HRNetV2-W28, pre-trained fully supervised; as it typically leads to higher $P_s$).

\begin{figure}[t]
\vspace{-.8cm}
  \includegraphics[width=8cm]{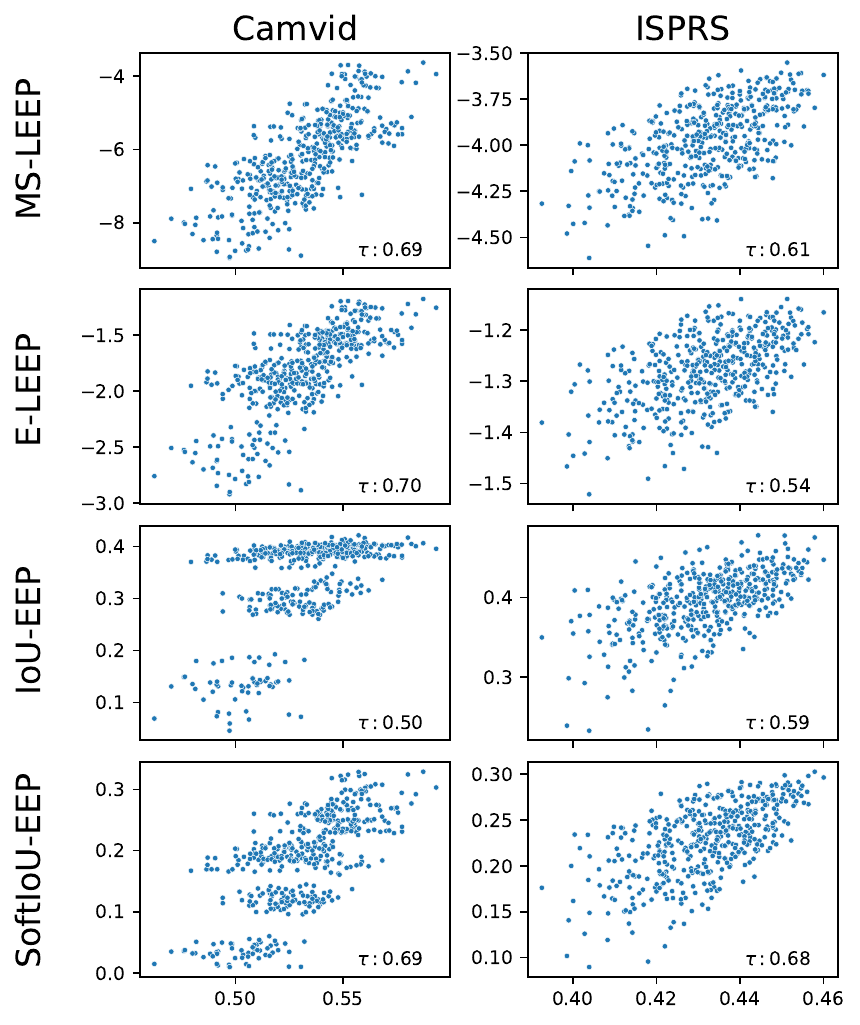}
  \centering
   \caption{
   Relation between the predicted transferability (y-axis) and the actual mean IoU performance (x-axis) on the target test set (for 2 of the 5 target datasets). Each plot shows 455 candidate ensembles as a separate dot, and also reports the corresponding weighted Kendall's $\tau$ correlation score.
   These scores are generally high, demonstrating the success of our transferability metrics.}
   \label{fig:scatterplot_transferability_metrics}
   \vspace{-.4cm}
\end{figure}

\subsection{Results} \label{correlation-results}

\cref{fig:scatterplot_transferability_metrics} shows qualitatively the relation between predicted transferability and actual mean IoU for all our transferability metrics on two target datasets.
In all experiments we see good positive correlations, demonstrating that our metrics work well.
We also see that ensemble performance varies greatly depending on the sources used, justifying the importance of having a good ensemble selection mechanism.

\begin{figure}[t]
\vspace{-.7cm}
  \includegraphics[width=7.0cm]{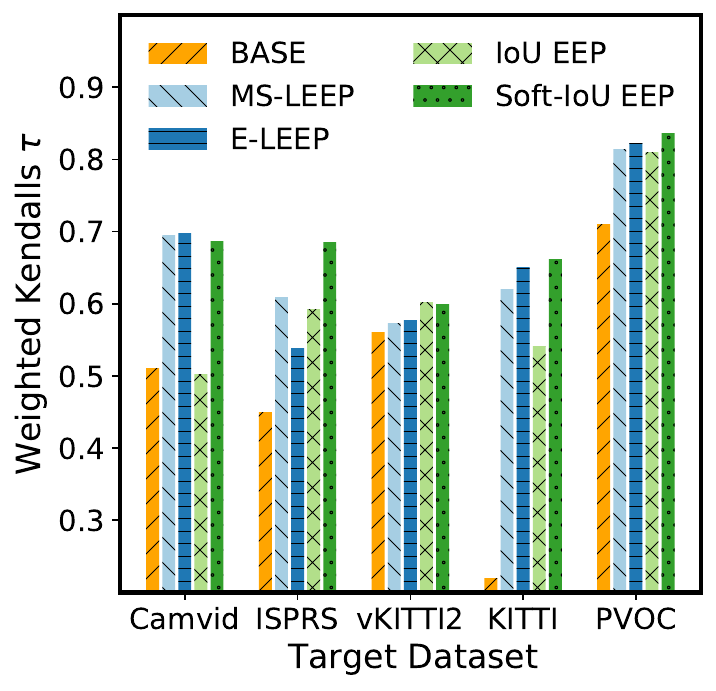}
  \centering
   \caption{Comparison of our transferability metrics, over 5 target datasets. We show the weighted Kendall's $\tau$ between a metric and actual mean IoU on the target test set, where each ranks all 455 candidate ensembles. \softiou{} performs best overall.}
   \label{fig:comparison_transferability_metrics}
   \vspace{-.4cm}
\end{figure}

Quantitatively, \cref{fig:comparison_transferability_metrics} reports the weighted Kendall's $\tau$
for our transferability metrics and the baseline (for all 5 target datasets). All our transferability metrics generally achieve high scores, significantly outperforming the baseline on each target dataset.
Among our metrics, the direct LEEP variants, MS-LEEP and E-LEEP, perform equally well.
Next, IoU-EEP performs the worst, suggesting that it is important to consider the probability distribution $P_{ens}(\cdot|x_i)$.
Finally, \softiou{} performs best, confirming the benefits of directly approximating the performance measure of the target test set (mean IoU).
On average, \softiou{} achieves $\tau$ of 69.3\%, outperforming the baseline by 20.3\%, \msleep{} by 3.1\%, \eleep{} by 3.6\%, \ioueep{} by 8.4\%.

As related works on single-source transferability metrics sometimes report the standard Kendall $\tau$~\cite{li21cvpr} or the (linear) Pearson coefficient~\cite{tan21cvpr,nguyen20icml, li21cvpr}, we also report them in \cref{tab:correlation_measures}. Despite each correlation measure is based on different mathematical assumptions, the general trend among target datasets and transferability metrics remains the same.

\begin{table*}
\vspace{-.8cm}
\small
  \centering
  \begin{tabular}{c|*{5}{@{\hspace{3.5mm}}c}|*{5}{@{\hspace{3.5mm}}c}|*{5}{@{\hspace{3.5mm}}c}}
    \toprule
    Measure & \multicolumn{5}{c}{Weighted Kendall's $\tau$} & \multicolumn{5}{c}{Kendall's $\tau$} & \multicolumn{5}{c}{Pearson} \\
    Method & BASE & MS & E & IoU & sIoU & BASE & MS & E & IoU & sIoU & BASE & MS & E & IoU & sIoU\\
    \midrule
    Camvid & 0.51 & 0.69 & \textbf{0.70} & 0.50 & 0.68 & 0.30 & \textbf{0.56} & 0.55 & 0.37 & 0.51 & 0.46 & \textbf{ 0.74} & 0.73 & 0.56 & 0.70\\
    ISPRS & 0.45 & 0.61 & 0.54 & 0.59 & \textbf{0.68} & 0.22 & 0.41 & 0.37 & 0.41 & \textbf{0.44} & 0.29 & 0.60 & 0.55 & 0.59 & \textbf{0.63}\\
    vKITTI2 & 0.56 & 0.57 & 0.58 & \textbf{0.60} & \textbf{0.60} & 0.38 & 0.49 & 0.51 & 0.51 & \textbf{0.53} & 0.54 & 0.66 & 0.67 & 0.62 & \textbf{0.70}\\
    KITTI & 0.22 & 0.62 & 0.65 & 0.54 & \textbf{0.66} & 0.17 & 0.50 & \textbf{0.54} & 0.42 & 0.53 & 0.25 & 0.69 & 0.73 & 0.62 & \textbf{0.74}\\
    PVOC  & 0.71 & 0.81 & 0.82 & 0.81 & \textbf{0.83} & 0.42 & 0.59 &\textbf{ 0.61} & 0.58 & \textbf{0.61} & 0.66 & \textbf{ 0.84} & 0.83 & 0.73 & 0.81 \\
    \bottomrule
    Average & 0.49 & 0.66 & 0.66 & 0.61 & \textbf{0.69} & 0.30 & 0.51 & \textbf{0.52} & 0.46 & \textbf{0.52} & 0.44 & 0.71 & 0.70 & 0.62 & \textbf{0.72}\\
  \end{tabular}
  \caption{Correlation measures for 5 different target datasets, obtained by comparing the transferability predicted by a metric to the actual Mean IoU on the target test set. The methods shortcuts refer to: (BASE=Baseline), (MS=\msleep{}), (E=\eleep{}), (IoU=\ioueep{}), (sIoU=\softiou{}). \softiou{} performs best overall. See Sec. \ref{multi-source-exp-setup} for more details about the setup.}
  \label{tab:correlation_measures}
  \vspace{-.4cm}
\end{table*}

\section{Evaluating ensemble selection} \label{sec:ensemble_selection}


\subsection{Experimental Setup} \label{sec:ensemble_selection_setup}

We now turn to ensemble selection: we use our best transferability metric, \softiou{}, to select the ensemble with the highest predicted transferability. We fine-tune this ensemble on the target training set and evaluate on the target test set.
Selecting only a single ensemble per target dataset on which to perform expensive fine-tuning and evaluation enables us to expand our experimental setup beyond what we did in Sec.~\ref{sec:evaluate-metrics}. Instead of using $N=15$, we now use all $N=64$ source models as candidates. Moreover, instead of only fine-tuning individual ensemble members, we add an additional step of ensemble-specific fine-tuning.
We compare to two baselines which select a single source model.


\para{Improved actual performance of ensembles.}
Given an ensemble, we start by fine-tuning each member individually on the target training set as in Sec.~\ref{multi-source-exp-setup}.
Afterwards, we perform additional ensemble-specific fine-tuning to improve it by re-weighting the class predictions of its members.
Specifically, we freeze the backbone of each member and attach to it a light head that assigns per-class weights and biases, before averaging their predictions into a single ensemble prediction.
We then fine-tune the light head of this ensemble on the target training set.
While the total number of additional parameters introduced by this head depends on the number of target classes, in practice it is between 144 and 2700, which is negligible compared to the total number of parameters in the models composing the ensemble.
 

\para{Ensemble size.} We set $S=3$ as in Sec.~\ref{sec:evaluate-metrics}. 

\para{Target datasets.} 
We now consider each of the 17 datasets in~\cref{tab:datasets} as a target dataset in turn (instead of 5 in Sec. \ref{sec:evaluate-metrics}).
As in Sec.~\ref{sec:evaluate-metrics}, we study transfer learning in the low data regime defined by~\cite{mensink21arxiv} (i.e. 150 target training images for each dataset, except 1000 for COCO and ADE20k as they contain many classes).

\para{Baseline B1: select a single source model.}
We test whether aggregating multiple source models is beneficial to transfer learning, compared to the standard of selecting a single source model~\cite{nguyen20icml, li21cvpr, tan21cvpr, you21icml}.
To do so, for a given target dataset we use the same pool of 64 source models as for ensemble selection. Then we select the single model with the highest predicted transferabiliy according to LEEP~\cite{nguyen20icml}. Finally, we fine-tune it on the target training set, and evaluate on the target test set.

\para{Baseline B1L: select a single \emph{large} source model.} \label{para:single-model-finetuning}
We also propose a stronger baseline which selects a single large model with the same number of parameters as one of our ensembles. An ensemble of $S=3$ has 69M parameters, as both HRNetV2-W28 and ResNet23M have 23M parameters.
We use the best such model according to preliminary experiments: HRNetV2-W48 pre-trained fully supervised on ILSVRC'12.
It offers excellent performance for semantic segmentation~\cite{wang20pami} and was also used in~\cite{lambert20cvpr, mensink21arxiv}.

For this baseline we need to construct a new pool of large source models. To do so, we train 17 large models, one for each source dataset (Tab. \ref{tab:datasets}). Then, given a target dataset, we use LEEP~\cite{nguyen20icml} to select a single model (out of 16, excluding that target dataset).

\subsection{Results} \label{sec:ensemble_selection_results}

\begin{figure}[t!]
  \vspace{-.1cm}
  \includegraphics[width=8cm]{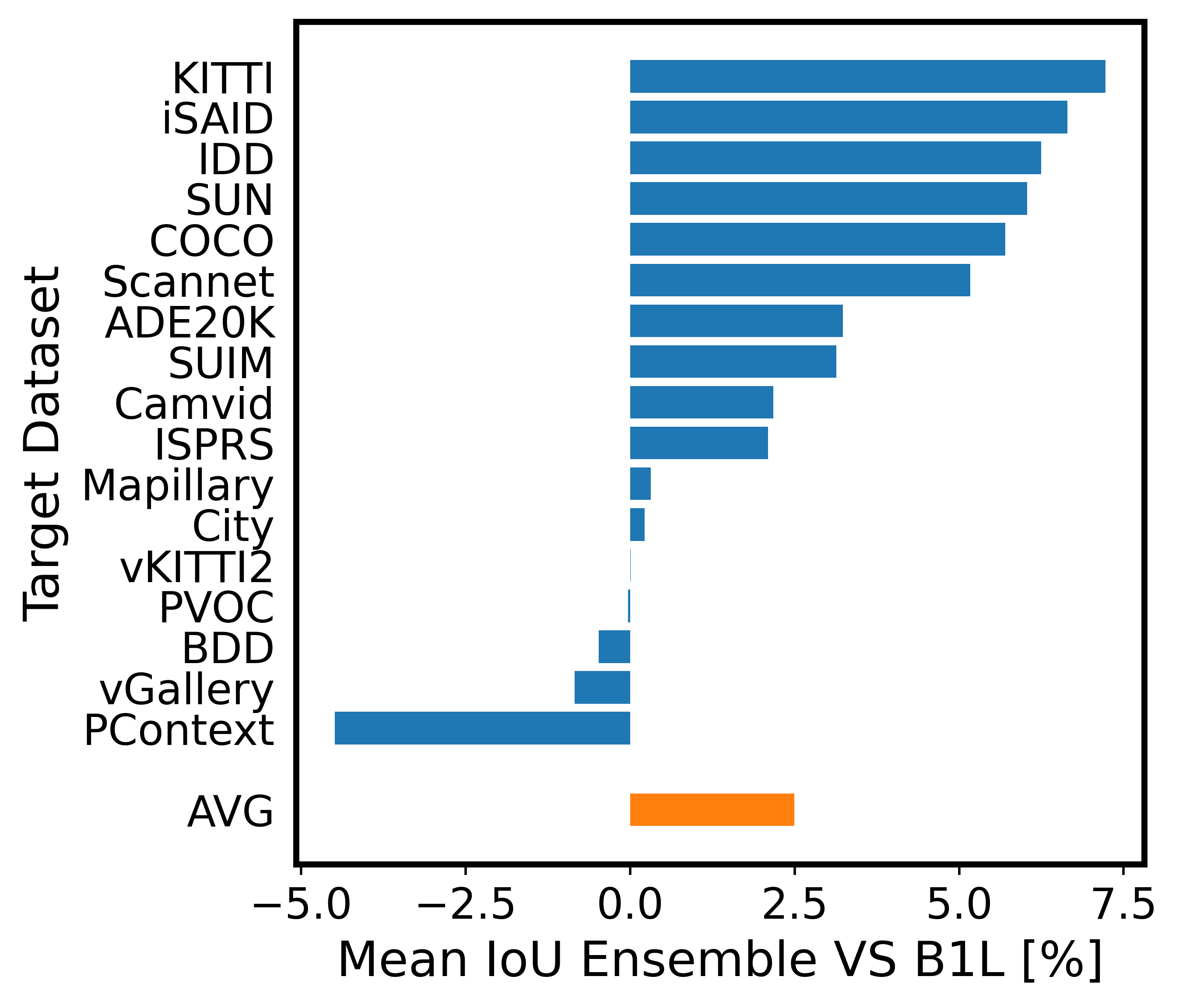}
  \centering
   \caption{Relative mean IoU gain on each target test set made by the ensemble selected by our \softiou{} metric over baseline B1L.
   For clarity, we set the 0 vertical line to corresponds to the performance of B1L. On average over all target datasets, the improvement is 2.5\%.
   We do not explicitly display baseline B1, as it performs worse than B1L (see main text for discussion).
   }
   \label{fig:ensemble_vs_baseline_relative_gain}
   \vspace{-.4cm}
\end{figure}

\begin{figure*}[t]
    \vspace{-.8cm}
  \includegraphics[width=16cm]{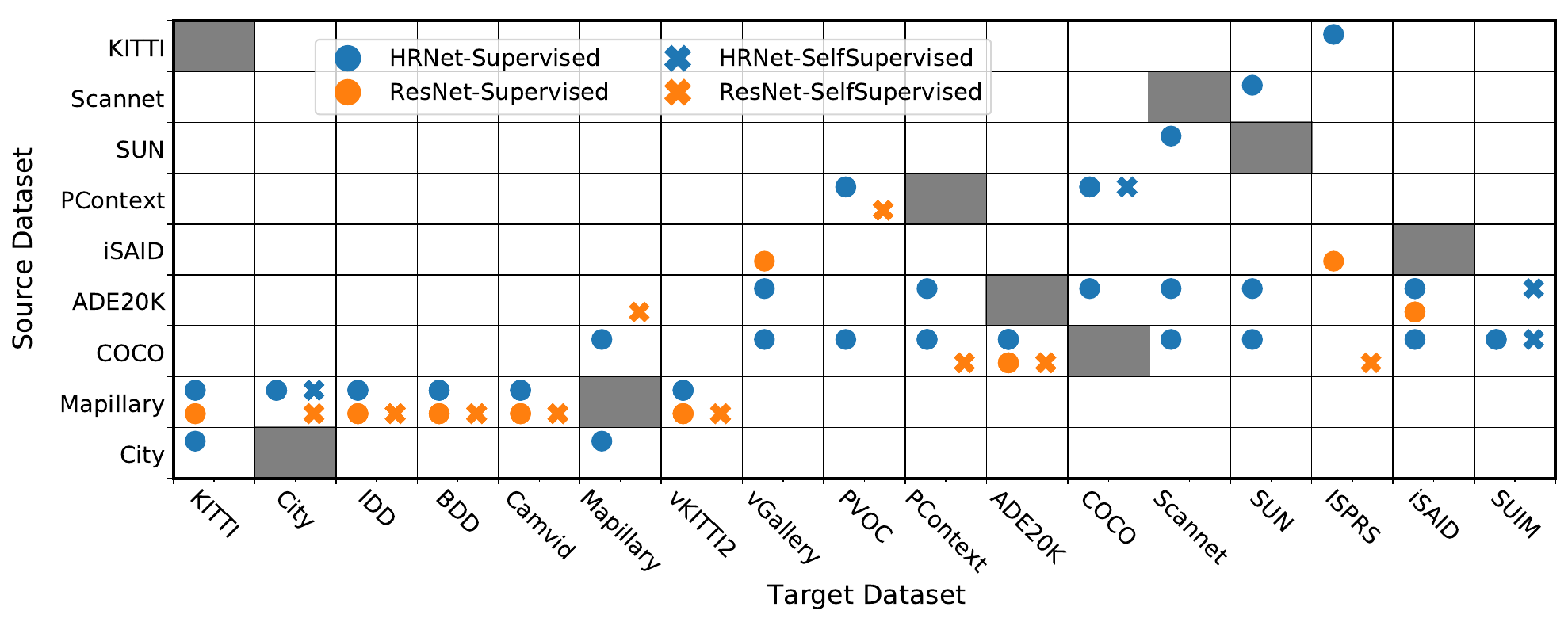}
  \centering
   \caption{For each target dataset (x-axis), we show the 3 source models in the ensemble selected by \softiou{}. When multiple source datasets cover the image domain of the target, our metric generates diversity by selecting models trained on these different source datasets (see target ScanNet and Sun). When there exists a single strong source dataset for a target, our metric selects only this source dataset, and instead generates diversity by varying architectures and pre-training schemes (see targets from the driving domain).
   Note: we do not consider source models trained on the same dataset as the target (gray cells).
   }
   \vspace{-.3cm}
   \label{fig:soft_iou_selected_ensembles}
\end{figure*}

\para{Comparison to baselines B1 and B1L.}
We first consider the improvement made by the ensemble selected by our transferability metric \softiou{} over the baseline B1.
On average across all 17 dataset, selecting an ensemble instead of a single model out of the same pool (B1) improves results by a relative +6.0\% mean IoU. This demonstrates that the source models in our pool are complementary and composing several into an ensemble brings clear benefits.

Next, we show in Fig.~\ref{fig:ensemble_vs_baseline_relative_gain} that
selecting an ensemble out of our pool even outperforms baseline B1L, which selects a single model out of a pool of {\em larger models}, each with capacity similar to one of our ensembles (+2.5\% relative mean IoU).
When looking at individual target datasets,
the selected ensemble improves over B1L by more than 2\% in 10 of them.
In contrast, B1L outperforms the ensemble only once (on Pascal Context).
We conclude that selecting an ensemble brings solid accuracy benefits over selecting a single model, even when controlling for total capacity.

\para{What factors of diversity matter more?}
Generally, ensembles benefit from the predictive diversity of its member models \cite{bian21ieee, dietterich00springer, lu10sigkdd, guo18neurocomp}.
To understand which factors of diversity are important in our transfer learning setting (Sec. \ref{preparing_source_models}), \cref{fig:soft_iou_selected_ensembles} shows which source models are selected by \softiou{} as part of the winning ensemble for each target dataset.

We observe the following patterns:
(1) Some target datasets are well covered by multiple source datasets. For these cases, our metric generates diversity by selecting source models trained on these different datasets. For example, for ScanNet (indoor) as a target, our metric selects source models trained on Sun (indoor), COCO, and ADE20K, as COCO and ADE20K contain many indoor images.
In these cases the backbone of choice is HRNetV2-W28 using fully supervised pre-training, which is generally the best perfoming backbone (76\% of all source models selected).
(2) For other target datasets there exists a strong source dataset which alone already covers most variations in the target domain. For example, Mapillary is the largest dataset in the driving domain and covers all continents. Our method nicely picks Mapillary almost exclusively as the sole source dataset for all target driving datasets (Cityscapes, IDD, BDD, Camvid, vKITTI2, and KITTI as the only case with a second source - Cityscapes).
In these cases, our metric generates diversity by varying model architectures and pre-training schemes within the selected ensemble.
(3) The most frequently selected source datasets have a greater number of training samples, greater number of labels, and larger diversity of images (Mapillary, COCO, ADE20K).
These observations are in line with earlier work which show the benefits of in-domain source images~\cite{ngiam18arxiv, mensink21arxiv, yan20cvpr} and large, broad source sets~\cite{kolesnikov20eccv, mahajan18eccv, mensink21arxiv, sun17iccv}.

\begin{figure}[t]
  \includegraphics[width=8cm]{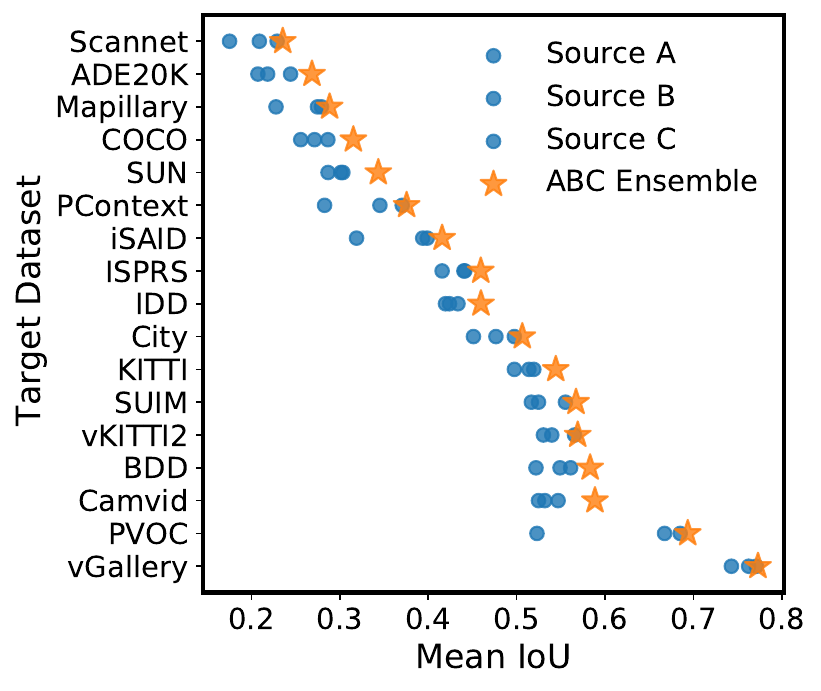}
  \centering
   \caption{Comparison of the mean IoU of the winning ensemble vs its component source models.}
   \label{fig:single_sources_vs_ensemble_eleep}
   \vspace{-.3cm}
\end{figure}

\para{Ensemble vs. its members.}
In \cref{fig:single_sources_vs_ensemble_eleep} we compare the mean IoU of the selected ensemble to that of its member models. On average the ensemble improves over its best member by 4.6\%, and over its worst member by 18.6\%.
These improvements demonstrate that the sources selected to be part of the winning ensemble are indeed diverse (otherwise the ensemble could not outperform its best member).
Hence we conclude that our transferability metric is good in selecting a diverse set of source models.

Finally, to evaluate how important it is to \emph{learn} to combine model predictions, we re-evaluate the results but now using a simple unweighted average of the model predictions instead of the learned head. This still outperforms the best single ensemble member by +2.9\%.  
Hence, our improvements are truly due to ensembling models (be it with a fixed combiner head, or with a learned one).

\section{Conclusion}

We design for the first time transferability metrics for ensemble selection.
We evaluate them in a challenging and realistic transfer learning setup for semantic segmentation, featuring 17 source datasets covering a wide variety of image domain, two model architectures, and two pre-training schemes.
We show experimentally that our transferability metrics rank correlate well with actual transfer learning performance. Moreover, our best metric selects an ensemble performing better than two baselines which select a single source model (even after equalizing capacity).

\appendix

\section{Limitations.}

    Here we discuss some limitations of our proposed method and of the general transfer learning field.
    
\begin{figure*}[b!]
\vspace{3cm}
\centering
\includegraphics[width=0.99\textwidth]{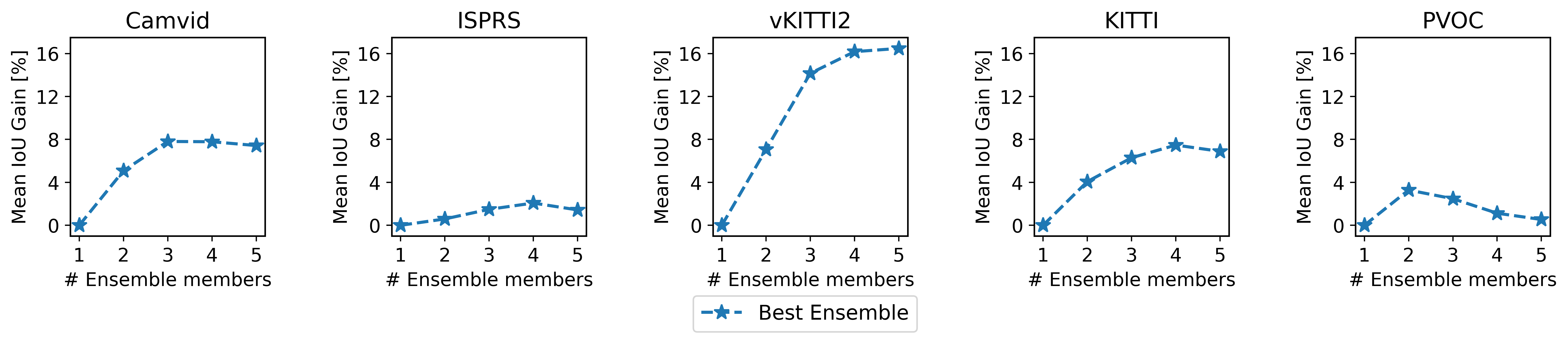}
\caption{Relative performance gain (Mean IoU) over the best single source model, of the best ensemble across $\binom{S}{15}$ possible combinations of $S$ ensemble members. We use a pool of 15 source models for each target dataset. We consider $S$ ranging from 1 to 5. Most performance gain comes from combining 3 source models together.}
\label{fig:ensemble_up_5_members}
\end{figure*}

\paragraph{Diverse Pool of Source Models.}
As discussed in \cref{sec:related-works}, ensembling machine learning models is a classical method for increasing accuracy, where having diverse models is typically important~\cite{breiman96ml,dietterich00springer,freund96icml,hansen90pami,krogh95nips,lee15arxiv}. Hence, the benefits of selecting an ensemble of source models over a single source model could decrease if the pool of source models is not diverse enough.

\balance

\paragraph{Focus on Limited Training Data.}
The general field of transfer learning focuses on the scenario of having a limited target training data, which is not sufficient for the model to achieve good performance and generalize to the target test data. Hence, by exploiting the knowledge of one or more source models, performance can be improved. For this reason, advances on this line of work may not be beneficial to scenarios where the target training data is enough for solving the final target test task.

\paragraph{General risks of transfer learning.}
As described in detail in \cite{bommasani21arxiv}, transfer learning demands caution, especially when using source models trained on broad data. In particular, problems related to the source models, such as intrinsic biases, can easily propagate to the target task.

\section{Training source models.}
We describe here implementation details on the training procedure of our source models.
We train using a pixel-wise cross-entropy loss, optimized by SGD with momentum \cite{qian99nn}.
We use 8 Google Cloud TPUs v3 with synchronized batch norm and batch size 32.
We decrease the learning rate by $10\times$ after $2/3$ of the total training steps.
For each model, we tune the initial learning rate and number of training steps on the source dataset.

\section{Ablation: number of ensemble members.}

We perform an ablation study to understand to which extend an ensemble can benefit from combining more source models together, given our pool of source models. We use ensembles composed of up to 5 members. We measure the actual performance of each ensemble using the target datasets and source models from \cref{sec:evaluate-metrics}. For example, when evaluating ensembles of 5 members out of a pool of 15 source models, we consider a total of $ \binom{15}{5} = 3003$ combinations. To limit computation, the source models are aggregated using an unweighted average of their predictions.

We are interested at the best performing ensemble, which is the one a transferability metric aims at selecting. Hence, we show in \cref{fig:ensemble_up_5_members} the performance of the best ensemble across all combinations for a given number of ensemble members. In detail, we show the relative performance gain compared to the best single source model (ensemble of 1 member).
Most performance gain comes from combining 3 models. This validates our choice to use 3 source models as compromise between performance and total capacity, as done in \cref{sec:evaluate-metrics} and \cref{sec:ensemble_selection}.

\clearpage
{\small
\bibliographystyle{ieee_fullname}
\bibliography{shortstrings,loco,loco_extra}
}

\end{document}